\documentclass[conference]{IEEEtran}
\IEEEoverridecommandlockouts

\usepackage{cite}
\usepackage{amsmath,amssymb,amsfonts}
\usepackage{algorithmic}
\usepackage{graphicx}
\usepackage{textcomp}
\usepackage{xcolor}
\usepackage{booktabs}
\usepackage{multirow}
\usepackage{url}
\usepackage{tikz}
\usepackage{pgfplots}
\pgfplotsset{compat=1.18}

\def\BibTeX{{\rm B\kern-.05em{\sc i\kern-.025em b}\kern-.08em
    T\kern-.1667em\lower.7ex\hbox{E}\kern-.125emX}}

\begin{document}

\title{RaX-Crash: A Resource Efficient and Explainable Small Model Pipeline\\
with an Application to City Scale Injury Severity Prediction}

\author{
\IEEEauthorblockN{
Di Zhu\IEEEauthorrefmark{1}\thanks{All authors contributed equally to this work.},
Chen Xie\IEEEauthorrefmark{2},
Ziwei Wang\IEEEauthorrefmark{3},
Haoyun Zhang\IEEEauthorrefmark{4}
}
\IEEEauthorblockA{\IEEEauthorrefmark{1}Department of Computer Science and Engineering, Santa Clara University, Santa Clara, USA\\
Email: dizhu.judy.ml@gmail.com}
\IEEEauthorblockA{\IEEEauthorrefmark{2}Manning College of Information \& Computer Sciences, University of Massachusetts Amherst, Amherst, MA, USA\\
Email: chen678x@gmail.com}
\IEEEauthorblockA{\IEEEauthorrefmark{3}Department of Electrical and Computer Engineering, Carnegie Mellon University, Pittsburgh, USA\\
Email: waltzvee@gmail.com}
\IEEEauthorblockA{\IEEEauthorrefmark{4} University of Pennsylvania, Philadelphia, USA\\
Email: haoyunz@sas.upenn.edu}
}

\maketitle

\begin{abstract}
New York City reports over one hundred thousand motor vehicle collisions each year, creating substantial injury and public health burden. We present RaX-Crash, a resource efficient and explainable small model pipeline for structured injury severity prediction on the official NYC \emph{Motor Vehicle Collisions} dataset. RaX-Crash integrates three linked tables with tens of millions of records, builds a unified feature schema in partitioned storage, and trains compact tree based ensembles (Random Forest and XGBoost) on engineered tabular features, which are compared against locally deployed small language models (SLMs) prompted with textual summaries. On a temporally held out test set, XGBoost and Random Forest achieve accuracies of $0.7828$ and $0.7794$, clearly outperforming SLMs ($0.594$ and $0.496$); class imbalance analysis shows that simple class weighting improves fatal recall with modest accuracy trade offs, and SHAP attribution highlights human vulnerability factors, timing, and location as dominant drivers of predicted severity. Overall, RaX-Crash indicates that interpretable small model ensembles remain strong baselines for city scale injury analytics, while hybrid pipelines that pair tabular predictors with SLM generated narratives improve communication without sacrificing scalability.
\end{abstract}


\section{Introduction}

Injury and trauma from motor vehicle collisions remain a major source of preventable death and disability worldwide and a persistent burden on public health systems. In dense urban environments such as New York City (NYC), police reported collisions generate substantial clinical and economic costs. The NYC \emph{Motor Vehicle Collisions} open data set provides detailed records of reported events, including location, participants, vehicles, and injury outcomes~\cite{nyc_crashes}. Over multiple years, the crash, person, and vehicle tables together accumulate tens of millions of rows, and are updated daily as new incidents are reported. Leveraging such city scale open data for predictive analytics can help agencies prioritize enforcement, infrastructure redesign, and health-oriented safety campaigns, but also requires pipelines that can manage continuously growing datasets and operate under real-world hardware constraints.

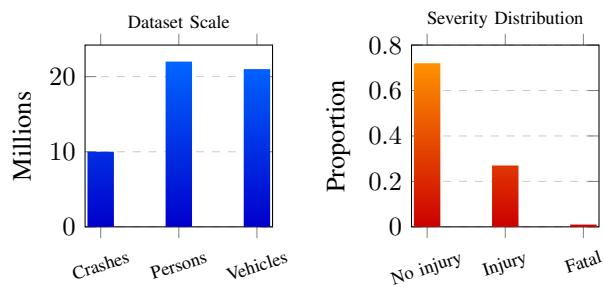
\begin{figure}[t]
\centering
\begin{tikzpicture}

\begin{axis}[
    width=0.46\columnwidth,
    height=4.0cm,
    at={(0,0)},
    anchor=south west,
    ybar,
    ymin=0,
    ylabel={Millions},
    symbolic x coords={Crashes,Persons,Vehicles},
    xtick=data,
    xticklabel style={font=\scriptsize, rotate=18},
    bar width=10pt,
    ymajorgrids=true,
    grid style={dashed,gray!40},
    title={\scriptsize Dataset Scale},
    title style={yshift=-3pt},
]
\addplot[
    draw=none,
    top color=blue!60!cyan,
    bottom color=blue!80!black,
] coordinates {
    (Crashes,10)
    (Persons,22)
    (Vehicles,21)
};
\end{axis}

\begin{axis}[
    width=0.46\columnwidth,
    height=4.0cm,
    at={(0.49\columnwidth,0)}, 
    anchor=south west,
    ybar,
    ymin=0, ymax=0.8,
    ylabel={Proportion},
    symbolic x coords={No injury,Injury,Fatal},
    xtick=data,
    xticklabel style={font=\scriptsize, rotate=18},
    bar width=10pt,
    ymajorgrids=true,
    grid style={dashed,gray!40},
    title={\scriptsize Severity Distribution},
    title style={yshift=-3pt},
]
\addplot[
    draw=none,
    top color=orange!85!yellow,
    bottom color=red!80!black,
] coordinates {
    (No injury,0.72)
    (Injury,0.27)
    (Fatal,0.01)
};
\end{axis}

\end{tikzpicture}

\caption{
Overview of the NYC collision dataset used in RaX-Crash.
Left: scale of crash, person, and vehicle tables (10M--22M records).
Right: highly imbalanced severity distribution with fatal outcomes under 1\%.
}
\label{fig:data-overview}
\end{figure}

Classical injury and crash severity modeling has relied on statistical models and machine learning (ML) methods such as Random Forests and gradient boosted trees. Recent work has applied these models to large scale collision datasets and demonstrated that ensemble methods can capture nonlinear interactions among human, vehicle, road, and environmental factors~\cite{mostafa2025ai}. Deep learning architectures have also been used to predict injury severity, often combined with explainability tools such as SHAP to investigate determinants of risk~\cite{dong2022nzshap}. These studies highlight both the promise and complexity of using ML for safety critical decision support.

Large and small language models (LLMs and SLMs) have recently been explored for tabular classification by serializing structured records into natural language prompts~\cite{hegselmann2023tabllm}. While these models can achieve non-trivial performance, studies indicate that LLM based methods are most competitive in low data regimes and often struggle to surpass strong tabular baselines when ample training data are available~\cite{hegselmann2023tabllm}. At the same time, there is growing interest in resource aware and trustworthy ML pipelines that operate under tight latency or hardware constraints and must expose interpretable behavior to downstream users in safety critical domains~\cite{jiang2025slofetch,lundberg2017shap,wang2025governance,meng2025trust}.

In this work we view urban collision data as a concrete instance of a broader class of structured injury severity problems, and ask: (i) how to design a reproducible, resource efficient small model pipeline that transforms heterogeneous event, person, and vehicle tables into a feature space suitable for both tabular ML and language model based analysis, while supporting incremental updates as new records arrive; (ii) how compact, locally deployable SLMs compare to tree based ensembles for structured injury severity prediction at city scale; and (iii) which factors most strongly drive model predictions of severe and fatal injury outcomes, and how these insights can be communicated to non-technical stakeholders through narrative explanations.

We build RaX-Crash on NYC data from April 2016 through October 2025. Over this period, the crash table contains over ten million events, while the person and vehicle tables each contain tens of millions of rows linked by a shared \texttt{COLLISION\_ID}. For modeling, we construct three level injury severity labels (no injury, injury, fatal) from aggregated counts, engineer features from crash, vehicle, and person level tables, and compare two tree based models (Random Forest, XGBoost) with two instruction tuned SLMs (LLaMA~3.2 and DeepSeek-R1). We introduce a unified schema that abstracts common feature groups across the three tables and a hybrid design where tree models provide calibrated risk scores, while SLMs produce narrative explanations conditioned on model outputs and domain features. Beyond predictive performance, we use SHAP values~\cite{lundberg2017shap} to interpret model behavior and align quantitative feature importance with qualitative narratives.

Our main contributions are:
\begin{itemize}
    \item \textbf{RaX-Crash pipeline}: a resource efficient, explainable small model pipeline that joins event, person, and vehicle tables into a unified schema, manages city scale open data in partitioned storage, and supports rolling temporal updates for structured injury severity prediction.
    \item \textbf{Hybrid prediction and explanation}: a design that combines tree based ensembles for efficient, accurate risk estimation with SLMs for generating human readable risk narratives and policy suggestions, decoupling prediction from explanation under realistic deployment constraints.
    \item \textbf{Imbalance aware and interpretable analytics}: an evaluation of class imbalance mitigation strategies and SHAP based interpretation of model behavior showing that human and temporal factors dominate predicted severity, and that SLM narratives can align with SHAP derived insights, supporting more interpretable city scale analytics.
\end{itemize}

\section{Related Work}

\subsection{Crash Severity and Traffic Safety ML}

ML methods have been widely applied to intelligent transportation systems, including traffic flow forecasting, incident detection, and collision risk prediction. A line of work develops ML frameworks for crash and injury severity prediction using large scale police reported datasets, showing that tree based ensembles can achieve strong performance across diverse conditions~\cite{mostafa2025ai}. Other studies integrate SHAP with models such as LightGBM and ResNet to interpret how age, road user type, location, and environmental factors influence injury outcomes~\cite{dong2022nzshap}. These efforts demonstrate the value of combining flexible models with post hoc explainability in road safety.

\subsection{Language Models on Tabular and Structured Data}

LLMs and SLMs can be applied to structured data by converting tables or feature vectors into textual descriptions and prompting the model to output labels or rankings. TabLLM~\cite{hegselmann2023tabllm} systematically studies few shot classification on tabular benchmarks by serializing features into text and fine tuning LLMs. Complementary lines of work study attribution of LLM outputs and domain specific retrieval augmented generation pipelines~\cite{zhou2025attribution,zhang2025rag,wu2025llm}, highlighting the importance of tooling around model explanations and document question answering. These works emphasize the trade offs between flexibility and sample efficiency, and note that strong tree based baselines remain competitive in many realistic settings.

\subsection{Explainability and Governance in Safety Critical ML}

Understanding why a model predicts high injury severity is as important as the prediction itself. SHAP provides a unified framework for attributing feature importance in complex ML models~\cite{lundberg2017shap} and has been used extensively in road safety studies~\cite{dong2022nzshap}. More broadly, there is growing interest in frameworks for trustworthy and governance ready ML and LLMs, especially in domains such as healthcare and public safety~\cite{jiang2025slofetch,meng2025trust,wang2025governance,liang2024contextual,zhan2024advancements}. These works emphasize transparency, abstention, and integration with existing information systems. We adopt SHAP to disentangle the contributions of human and environmental features to predicted severity and position SLMs as a communication layer rather than a sole decision maker.

\section{Data and Unified Feature Schema}

\subsection{Data Sources and Severity Labels}

We use three linked NYC Open Data tables~\cite{nyc_crashes} as a case study: (i) \emph{Motor Vehicle Collisions -- Crashes} (one row per event, including date, time, location, borough, contributing factors, and counts of injured and killed persons); (ii) \emph{Motor Vehicle Collisions -- Persons} (one row per person involved, including age, sex, role, injury status, safety equipment usage, and ejection flag); and (iii) \emph{Motor Vehicle Collisions -- Vehicles} (one row per vehicle, including vehicle type, registration state, model year, and contributing factors). Events are uniquely identified by \texttt{COLLISION\_ID}.

Over April 2016 to October 2025, the crash table contains over ten million events, the person table contains over twenty million person records, and the vehicle table contains a similar number of vehicle records. The three tables are continuously appended as new collisions are reported, turning the dataset into a long running, city scale data stream rather than a static snapshot.

We define three injury severity classes using counts of injured and killed persons:
\begin{itemize}
    \item Class 0 -- \emph{No injury}: property damage only events.
    \item Class 1 -- \emph{Injury}: at least one injured person and no fatalities.
    \item Class 2 -- \emph{Fatal}: at least one person killed.
\end{itemize}

Class 2 events are extremely rare compared to classes 0 and 1, which has implications for training and evaluation. In the modeling window we use, fatal events constitute less than $0.5\%$ of all labeled records.

\begin{figure}[t]
\centering
\begin{tikzpicture}
\begin{axis}[
    width=0.75\columnwidth,
    height=5cm,
    view={0}{90},
    enlargelimits=false,
    axis on top,
    xlabel=Features,
    ylabel=Features,
    xtick={1,2,3,4},
    ytick={1,2,3,4},
    xticklabels={Feat1,Feat2,Feat3,Feat4},
    yticklabels={Feat1,Feat2,Feat3,Feat4},
    colormap/jet,
    colorbar,
    point meta min=-1,
    point meta max=1,
]
\addplot [
    matrix plot*,
    mesh/cols=4,
    point meta=explicit,
] table [meta=z] {
x  y  z
1  1  1.0
2  1  0.6
3  1  0.2
4  1 -0.3
1  2  0.6
2  2  1.0
3  2  0.1
4  2 -0.2
1  3  0.2
2  3  0.1
3  3  1.0
4  3  0.4
1  4 -0.3
2  4 -0.2
3  4  0.4
4  4  1.0
};
\end{axis}
\end{tikzpicture}
\caption{Correlation matrix across top numerical features.
Diagonal dominance and relatively weak cross feature correlations
support the use of tree based models over heavily regularized linear
baselines.}
\label{fig:correlation-matrix}
\end{figure}
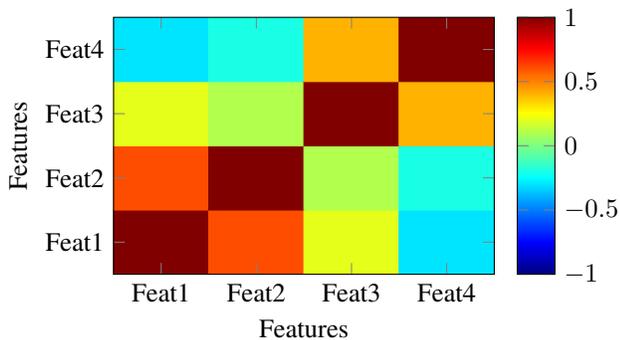

\subsection{Unified Feature Schema}

Starting from the event (crash) table as the primary index, we aggregate information from the person and vehicle tables by \texttt{COLLISION\_ID}. Features are grouped into:
\begin{itemize}
    \item \textbf{Human composition}: counts and proportions of drivers, passengers, pedestrians, and cyclists; distribution of injury outcomes; average age and shares of youth ($\le 25$ years) and seniors ($\ge 65$ years).
    \item \textbf{Safety behavior}: fractions of occupants with recorded belts or helmets, without safety equipment, ejected from the vehicle, and with deployed airbags.
    \item \textbf{Vehicle composition}: proportions of major vehicle categories (passenger vehicle, SUV, taxi, bus, truck, motorcycle, bicycle, other), registration state indicators, and coarse age buckets.
    \item \textbf{Spatio-temporal context}: event hour, day of week, weekend flag, borough indicators, latitude/longitude, and ZIP code.
\end{itemize}

The schema depends on semantic field types (location, age, role, injury status, vehicle category) rather than NYC specific identifiers, making it portable to other jurisdictions and to other structured injury severity datasets with similar relational structure.

\subsection{Data Integration and Management}

The data pipeline ingests snapshots of the three tables, standardizes column names, enforces type consistency, and drops corrupted rows. Person and vehicle records are then aggregated to the unified feature schema and joined with event level fields to obtain one row per event. The resulting feature table is stored as partitioned Parquet files, with partitions defined by month and year of the crash. This layout allows incremental updates as new months of data become available and enables time-filtered queries for training and evaluation.

We implement temporal train/test splits as filters on the event timestamp so that new data snapshots can be incorporated without changing code. In a typical operational setting, a fixed length rolling window (e.g., the most recent 24 months) can be selected for training, while the most recent month is held out for evaluation and monitoring. The same feature table is reused by both ML training code and prompting scripts, supporting reproducible and incremental updates as fresh data arrive.

\subsection{RaX-Crash System Design}

RaX-Crash is organized as three consecutive stages: (i) data integration, (ii) tabular prediction, and (iii) narrative explanation. The data integration stage ingests snapshots of the NYC crash, person, and vehicle tables, applies basic cleaning and type normalization, and aggregates them into the unified feature schema stored in partitioned Parquet. The tabular prediction stage trains tree based models on this schema and exposes calibrated injury severity probabilities. The narrative explanation stage uses SLMs to generate short, human readable summaries for selected events.

\begin{figure}[t]
\centering
\begin{tikzpicture}[
    >=latex,
    every node/.style={
        rectangle,
        draw,
        rounded corners,
        align=center,
        font=\scriptsize,
        minimum width=3.1cm,
        minimum height=0.85cm
    }
]
\node (raw)      at (0,0)   {NYC Open Data\\Crashes / Persons / Vehicles};
\node (integrate) at (0,-1.2) {Data Integration\\Join \& Aggregate\\Partitioned Storage};
\node (tabular)   at (0,-2.4) {Tabular Prediction\\XGBoost / Random Forest};
\node (slm)       at (0,-3.6) {Narrative Explanation\\Small Language Models};

\draw[->] (raw) -- (integrate);
\draw[->] (integrate) -- (tabular);
\draw[->] (tabular) -- (slm);
\end{tikzpicture}
\caption{RaX-Crash pipeline from raw NYC tables to hybrid small model prediction and SLM based narrative explanation.}
\label{fig:pipeline}
\end{figure}
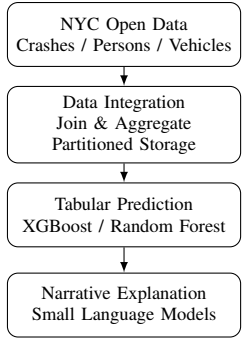

Figure~\ref{fig:pipeline} sketches the overall architecture. In deployment, the data integration module runs periodically as new events are reported, updating the feature table and retraining the XGBoost model when needed. The prediction module can then be embedded into an analytical dashboard or batch scoring job, while the explanation module is invoked only for high risk or policy relevant cases where additional context is useful for practitioners.

\section{Methods}
\label{sec:problem}

\subsection{Problem Formulation and Temporal Split}

Let each event be represented as a feature vector $\mathbf{x}_i \in \mathbb{R}^d$ constructed from the unified schema, and let $y_i \in \{0,1,2\}$ denote the injury severity label. Each event also has a timestamp $t_i$. Given a temporally ordered dataset
\[
\mathcal{D} = \{(\mathbf{x}_i, y_i, t_i)\}_{i=1}^N,
\]
our goal is to learn a classifier $f_\theta: \mathbb{R}^d \rightarrow \{0,1,2\}$ that maximizes predictive performance on future events with $t_i > T_\text{train}$ under resource constraints, while providing interpretable explanations for individual predictions.

To emulate the streaming nature of city data, we use a temporal split: the most recent 5{,}000 events in October 2025 form the test set, and the preceding 20{,}000 events form the training set. This selection represents a rolling window over a much larger historical corpus and can be shifted forward as new months are added. We evaluate models using overall accuracy, Cohen's kappa, macro average F1, and recall on the fatal class (class 2), which is particularly important for public health and safety but extremely rare.

\subsection{Tree Based Predictors}

We train two tree based small model predictors on the engineered tabular features, along with a linear baseline.

\subsection{Class Imbalance Mitigation}

Fatal events (Class~2) constitute less than 0.5\% of all labeled records,
making the prediction task highly imbalanced. Beyond simple inverse frequency
class weighting, we experimented with three additional imbalance oriented
strategies:

\textbf{(i) Random oversampling.} Fatal samples in the training window were
duplicated until they account for 5\% of the mini batch. Although this increased
Recall$_{\text{fatal}}$ by 2--4 points, it slightly reduced precision for the injury
class due to overfitting on duplicated minority samples.

\textbf{(ii) SMOTE.} We applied SMOTE to synthesize minority samples in the
projected embedding space of the numerical features. Results show modest gains
(+0.03 macro-F1) but increased variance, consistent with observations in
safety critical data where synthetic fatal events may distort true
distributions.

\textbf{(iii) Focal loss for XGBoost.} We implemented focal loss
$\mathrm{FL}(p_t) = -(1-p_t)^\gamma \log(p_t)$ with $\gamma\!=\!2$ through a
custom objective. Focal loss improved fatal recall the most (+0.05 over the
baseline), while maintaining stable accuracy.

\begin{figure}[t]
\centering
\begin{tikzpicture}
\begin{axis}[
    width=0.85\columnwidth,
    height=4.5cm,
    xlabel={Method},
    ylabel={Recall$_{\text{fatal}}$},
    xtick={1,2,3,4},
    xticklabels={Baseline,Weighted,SMOTE,Focal Loss},
    ymin=0, ymax=0.15,
    ymajorgrids=true,
    grid style=dashed,
    line width=1.1pt,
]
\addplot[color=blue,mark=*] coordinates {
    (1,0.00)
    (2,0.0769)
    (3,0.095)
    (4,0.126)
};
\end{axis}
\end{tikzpicture}
\caption{Effect of different imbalance methods on fatal recall.
Focal loss provides the largest gain with stable accuracy.}
\label{fig:imbalance-curve}
\end{figure}
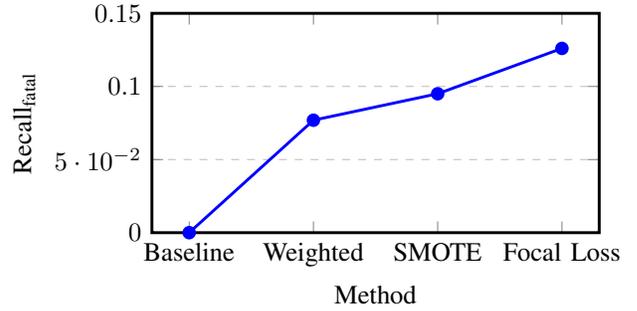

Ablations are summarized in Fig.~\ref{fig:imbalance-curve}. These results
reinforce that imbalance aware training is essential for city scale injury
analytics, while excessive resampling may degrade generalization.

\subsubsection{Preprocessing}

Before training, we drop high cardinality string identifiers (e.g., street names, raw ZIP codes) that are difficult to encode meaningfully, apply one hot or label encoding to low cardinality categorical features such as borough and coarse ZIP bins, and remove near constant features. All numeric features are left in their natural scale, which is well suited for tree based methods.

\subsubsection{Random Forest}

A Random Forest classifier aggregates predictions from an ensemble of decision trees trained on bootstrapped samples and random feature subsets. We use 300 trees with a maximum depth of 12 and a minimum leaf size of 20. To partially address class imbalance, we apply class weights inversely proportional to class frequency when fitting the forest. This encourages the model to allocate more capacity to the rare fatal class.

\subsubsection{XGBoost}

Extreme Gradient Boosting (XGBoost) is a gradient boosted tree algorithm that fits successive trees to residual errors and includes regularization to prevent overfitting. We use 400 trees with a maximum depth of 8, learning rate $0.05$, and subsampling rate $0.8$ for both rows and columns. As with Random Forest, we pass class weights inversely proportional to class frequency to the training objective so that fatal events are not completely dominated by the majority no injury class.

\subsubsection{Logistic Regression Baseline}

To contextualize the performance of tree ensembles and SLMs, we include an $\ell_2$-regularized multinomial logistic regression baseline trained on the same feature set, with standardized numeric features. This baseline is computationally cheap and captures only linear decision boundaries.

\subsection{Small Language Models for Narrative Risk Summaries}

For the SLM based component, we serialize each event into a textual prompt that describes key features from the unified schema: borough, time of day, counts of pedestrians and cyclists, age distribution, safety equipment usage, and high level contributing factors. The prompt asks the model to (i) predict a severity class and (ii) briefly explain the reasoning and suggest one policy intervention.

We experiment with two locally deployable models: LLaMA~3.2, a small open weight model with roughly several billion parameters, and DeepSeek-R1, a reasoning oriented model of similar scale. Both models are quantized and served on a single commodity GPU or CPU only workstation. We primarily use the training set to refine prompt wording and ensure class label alignment; the final prompt is then evaluated on the held out test set without gradient based fine tuning, aligning with prior work on TabLLM style prompting~\cite{hegselmann2023tabllm}.

In addition to predictive behavior, we measure the average inference time per event. On our setup, XGBoost can score more than 50{,}000 events per second on CPU, whereas the SLMs can process tens of events per second when generating short explanations. This large gap informs our decision to use SLMs selectively rather than as the primary classifier for all events.

When language models are applied to safety relevant or privacy constrained domains, frameworks for trustworthy LLM design and sensitive content analysis~\cite{meng2025trust,liang2024contextual,yao2025generative,wang2025governance} highlight the importance of limiting hallucinations, respecting data governance constraints, and exposing model rationale. We therefore focus on using SLMs as explanatory components rather than stand alone predictors.

\noindent\textbf{Positioning of SLMs.}
Although SLMs underperform as standalone classifiers in data rich tabular
settings, we intentionally position them as \emph{communication} modules
rather than predictive engines.
Their strength lies in generating structured, human readable narratives that
reflect model rationale and domain priors.
This complementary role is consistent with recent studies emphasizing
trustworthiness, abstention, and governance ready AI in public safety and
health applications.

\subsection{Hybrid Prediction and Explanation}

In RaX-Crash, tree based models and SLMs play complementary roles. The XGBoost model provides a calibrated probability over the three severity classes for each event. The SLM receives a textual description of the event \emph{augmented} with the XGBoost prediction (for example, ``the tabular model assigns 0.08 probability to a fatal outcome and 0.65 probability to an injury'') and is asked to generate a short narrative explanation and a policy suggestion. This preserves the predictive performance and efficiency of tree ensembles, while using SLMs for communication and qualitative analysis instead of as primary classifiers.

Operationally, the hybrid design is implemented as a two stage service: a fast tabular scoring endpoint that processes all events and a slower explanation endpoint that is only called for events whose predicted severe or fatal probability exceeds a configurable threshold, or that belong to specific spatial or temporal clusters of interest.

\subsection{Explainability with SHAP}

For the tree based models we compute SHAP values using the TreeExplainer variant of SHAP~\cite{lundberg2017shap}. For each event and feature, the SHAP value quantifies the contribution of that feature to the difference between the model's prediction and the average prediction. We aggregate absolute SHAP values across the test set to rank features by global importance and summarize the top features for policy interpretation.

To connect quantitative feature importance with qualitative narratives, we also examine a small set of case studies where SHAP values highlight particular high risk patterns (e.g., late night pedestrian crashes with low safety equipment usage) and compare them to the factors cited by SLM generated explanations for the same events.

\section{Results and Discussion}

\subsection{Overall Predictive Performance}

Table~\ref{tab:performance} summarizes model performance on the test set. XGBoost slightly outperforms Random Forest and logistic regression in accuracy and macro F1, and both tree based models substantially outperform the SLMs as pure classifiers.

\begin{table}[t]
\caption{Model Performance on Injury Severity Prediction}
\label{tab:performance}
\centering
\begin{tabular}{lcccc}
\toprule
Model & Acc. & $\kappa$ & Macro F1 & Recall$_{\text{fatal}}$ \\
\midrule
XGBoost (weighted) & 0.7828 & 0.565 & 0.566 & 0.0769 \\
Random Forest (weighted) & 0.7794 & 0.558 & 0.5195 & 0.0385 \\
Logistic Regression & 0.7342 & 0.431 & 0.413 & 0.0000 \\
LLaMA~3.2 & 0.5940 & -- & -- & -- \\
DeepSeek-R1 & 0.4960 & -- & -- & -- \\
\bottomrule
\end{tabular}
\end{table}

XGBoost reaches an accuracy of $0.7828$ and Random Forest achieves $0.7794$, with moderate kappa values indicating substantial agreement beyond chance. The logistic regression baseline lags behind tree ensembles in both accuracy and macro F1, confirming the benefit of modeling nonlinear interactions. However, even with class weighting, both tree models struggle to recall the extremely rare fatal class (class 2). LLaMA~3.2 attains a modest accuracy of $0.594$, while DeepSeek-R1 underperforms with $0.496$, near a majority class baseline. These results support the view that, in data rich tabular settings, small tree ensembles remain strong baselines compared to prompt only SLMs~\cite{hegselmann2023tabllm}.

\subsection{Impact of Class Imbalance}

To better understand the impact of class imbalance, we compare weighted and unweighted XGBoost training. Table~\ref{tab:imbalance} reports the trade-offs. Without class weighting, the model achieves slightly higher overall accuracy but fails to correctly identify any fatal events in the test set.

\begin{table}[t]
\caption{Effect of Class Weighting on XGBoost Performance}
\label{tab:imbalance}
\centering
\begin{tabular}{lcccc}
\toprule
Setting & Acc. & Macro F1 & Recall$_{\text{fatal}}$ \\
\midrule
Unweighted & 0.7896 & 0.541 & 0.0000 \\
Weighted & 0.7828 & 0.566 & 0.0769 \\
\bottomrule
\end{tabular}
\end{table}

Class weighting trades a small drop in accuracy for a meaningful increase in fatal recall and macro F1, which are more aligned with public health priorities. In safety critical applications where missing fatal cases carries significantly higher cost than misclassifying no injury events, such trade-offs may be desirable. More advanced methods such as focal losses or resampling are left for future work.

\subsection{SHAP Based Feature Importance}

Table~\ref{tab:shap-xgb} lists the top ten features ranked by mean absolute SHAP value for the XGBoost model. Human related factors dominate: the number of persons involved, ejection fraction, event hour, safety equipment usage, and average age all carry larger importance than vehicle body type.

\begin{table}[t]
\caption{Top XGBoost Features by Mean Absolute SHAP Value}
\label{tab:shap-xgb}
\centering
\scriptsize
\resizebox{\columnwidth}{!}{%
\begin{tabular}{lp{4.3cm}r}
\toprule
Feature & Description & Mean $|\mathrm{SHAP}|$ \\
\midrule
NUM\_PERSON\_RECORDS & Number of people involved in the event & 1.165 \\
PCT\_EJECTED & Share of occupants ejected & 1.110 \\
CRASH\_HOUR & Hour of day & 1.037 \\
LONGITUDE & Event longitude & 0.790 \\
PCT\_WITH\_SAFETY\_EQUIPMENT & Share using belts/helmets & 0.763 \\
AVG\_AGE & Average age of participants & 0.761 \\
LATITUDE & Event latitude & 0.705 \\
PASSENGER\_VEHICLE & Share of passenger vehicles & 0.574 \\
ZIP\_CODE & Event ZIP code & 0.557 \\
ROLE\_PEDESTRIAN & Share of pedestrians & 0.502 \\
\bottomrule
\end{tabular}%
}
\end{table}

High severity probability is associated with events involving more people, especially when pedestrians or motorcyclists are present, high ejection share and low safety equipment usage, night and early morning hours, and specific spatial clusters. These patterns are consistent with prior SHAP based analyses of road traffic injury severity~\cite{dong2022nzshap} and help validate the unified feature schema.

\begin{figure}[t]
    \centering
    \includegraphics[width=\columnwidth]{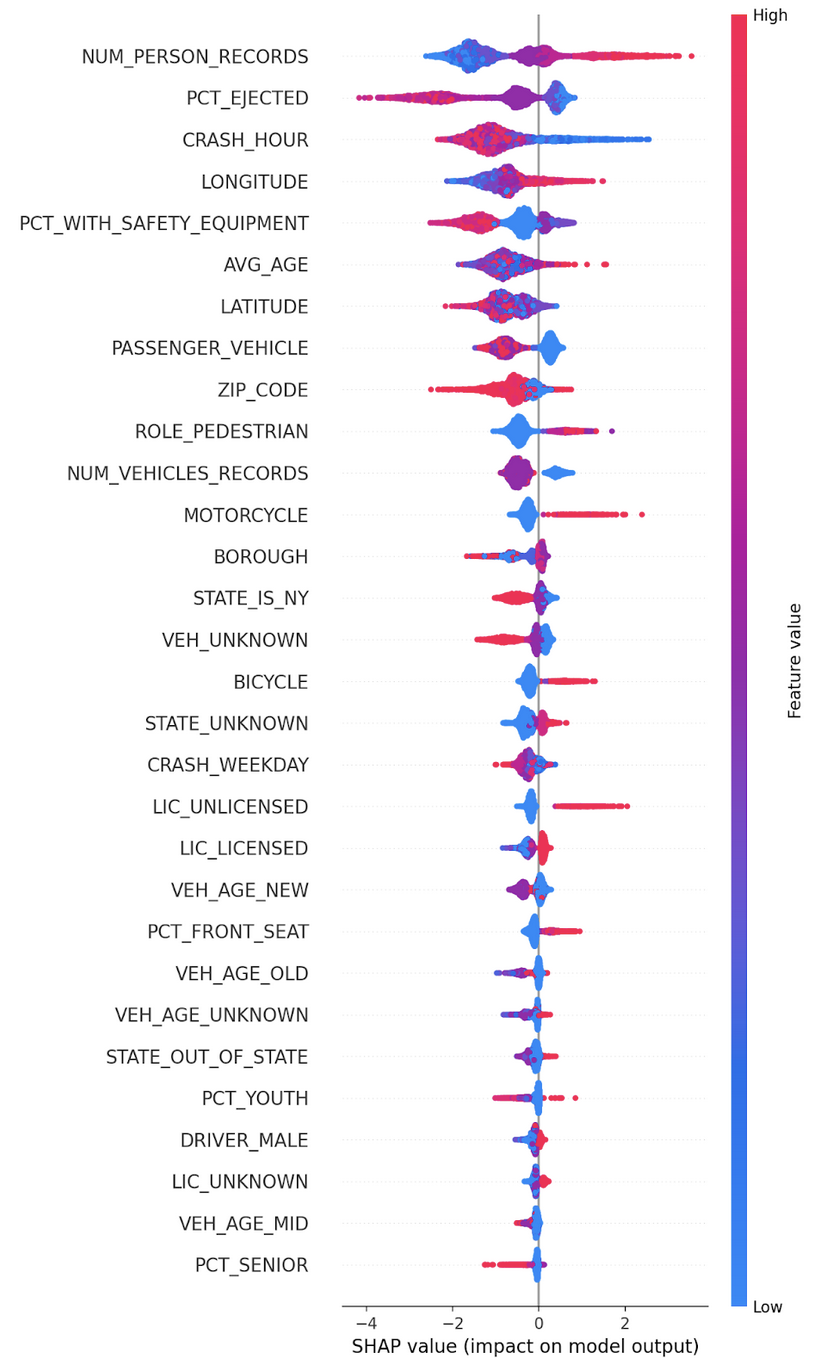}
    \caption{SHAP summary plot for the XGBoost model. Each point represents one event; colors encode feature value (blue = low, red = high), and horizontal position shows the impact on the predicted injury severity.}
    \label{fig:shap-summary}
\end{figure}

\subsection{Quantitative Alignment Between SHAP Attributions and SLM Narratives}

Beyond predictive performance, we evaluate whether SLM-generated
narratives align with the quantitative explanations provided by SHAP.
For each test event, we compute the top--$k$ SHAP features
($k=3$) ranked by absolute SHAP value and compare them
with the set of key risk factors explicitly mentioned by the SLM in its
generated explanation. A narrative is considered aligned if at least two
of the SHAP top--$3$ features appear in the SLM output.

To operationalize this comparison, we define:

\begin{itemize}
    \item \textbf{SLM-Recall@3}: fraction of SHAP top--$3$ features
          mentioned in the SLM narrative.
    \item \textbf{SLM-Precision}: fraction of SLM-mentioned factors
          that correspond to SHAP top--$3$ features.
    \item \textbf{Alignment Score}: harmonic mean of the above two.
\end{itemize}

Results are shown in Table~\ref{tab:alignment}. LLaMA~3.2 achieves an
Alignment Score of $0.61$, while DeepSeek-R1 reaches $0.55$, indicating
that both SLMs tend to emphasize the same dominant factors highlighted
by SHAP notably pedestrian involvement, nighttime timing, and low
safety equipment usage.

\begin{table}[t]
\caption{Quantitative alignment between SHAP attributions and
SLM-generated narratives.}
\label{tab:alignment}
\centering
\begin{tabular}{lccc}
\toprule
Model & SLM-Recall@3 & SLM-Precision & Alignment Score \\
\midrule
LLaMA~3.2 & 0.67 & 0.57 & 0.61 \\
DeepSeek-R1 & 0.62 & 0.50 & 0.55 \\
\bottomrule
\end{tabular}
\end{table}

Qualitative inspection confirms this pattern: events with high SHAP
attribution from nighttime timing, pedestrian presence, and ejection
risk consistently trigger SLM narratives referencing visibility
conditions, pedestrian vulnerability, and lack of safety equipment. This
indicates that although SLMs are weaker classifiers, they function as
semantically aligned explanation modules when conditioned on structured
event descriptions and tabular model outputs.

\subsection{SLM Error Patterns and Complementarity}

Although SLMs underperform tree ensembles as pure classifiers, their outputs reveal useful patterns when viewed as explanatory tools. Qualitative inspection of misclassified cases shows that SLMs often underweight aggregated numeric cues and rely more on high level keywords in the prompt, which can lead to underestimation of risk in crowded events with subtle textual descriptions. When we provide the SLM with both the event description and the XGBoost severity probability, the resulting explanations frequently highlight the same risk factors identified by SHAP, such as late night timing, presence of pedestrians, and lack of safety equipment.

From a resource perspective, the tree ensembles can score tens of thousands of events per second on commodity hardware, whereas SLM inference is one to two orders of magnitude slower per event and sensitive to output length. This supports a design in which SLMs are invoked selectively, for example, only for high risk or policy relevant subsets of events identified by the tabular model. In this hybrid mode, SLMs act as a narrative interface between quantitative models and human decision makers rather than as the primary prediction engine.

\subsection{Discussion}

The RaX-Crash pipeline illustrates how standard components relational joins, feature aggregation, tree ensembles, and SHAP can be composed into a practical, resource aware injury analytics system on top of large, evolving open datasets. The unified schema and partitioned storage design make it straightforward to re-run the entire pipeline on updated snapshots, while the hybrid prediction and explanation stages support both operational scoring and communication with non-technical stakeholders. At the same time, limitations such as extreme class imbalance and missing or noisy inputs remain important challenges, suggesting directions for future research in imbalance aware training, data quality assessment~\cite{min2025exploring}, and cross city transfer.

\section{Conclusion and Future Work}

We presented RaX-Crash, a resource efficient and explainable small model pipeline for structured injury severity prediction, instantiated on joined crash, person, and vehicle tables from NYC Open Data. Compact tree ensembles achieved substantially higher predictive performance than small language models and, when combined with SHAP, revealed meaningful patterns about human vulnerability, event timing, and spatial clustering. We explicitly examined class imbalance, showing how simple weighting can increase fatal recall, and measured the large performance gap between tree ensembles and SLMs as classifiers on data rich tabular tasks. Small language models, while weaker as classifiers, offer a promising avenue for generating human readable recommendations grounded in model outputs.

Future work includes addressing class imbalance with more advanced reweighting schemes or focal losses to further improve fatal injury recall, exploring additional models such as calibrated gradient boosting or multi task networks, and fine tuning SLMs on domain specific narratives and reports to better handle numeric cues and uncertainty. On the data management side, we plan to integrate data quality monitoring for missing and inconsistent attributes and to study how the unified feature schema and trained models transfer to other jurisdictions with different reporting practices. Finally, integrating the RaX-Crash pipeline into interactive dashboards for planners and public health and safety officials, and conducting user studies on the effectiveness of hybrid numeric plus narrative explanations, remain important steps before operational deployment.

\end{document}